\documentclass[pdflatex,sn-mathphys-num]{sn-jnl}% Math and Physical Sciences Numbered Reference Style 
\usepackage{graphicx}%
\usepackage{multirow}%
\usepackage{amsmath,amssymb,amsfonts}%
\usepackage{amsthm}%
\usepackage{mathrsfs}%
\usepackage[title]{appendix}%
\usepackage{xcolor}%
\usepackage{textcomp}%
\usepackage{manyfoot}%
\usepackage{booktabs}%
\usepackage{algorithm}%
\usepackage{algorithmicx}%
\usepackage{algpseudocode}%
\usepackage{listings}%

\theoremstyle{thmstyleone}%
\theoremstyle{thmstyletwo}%

\theoremstyle{thmstylethree}%

\begin{document}

\title[Article Title]{KRAIL: A Knowledge-Driven Framework for Base Human Reliability Analysis Integrating IDHEAS and Large Language Models}


\author[1]{ \sur{Xingyu Xiao}}\email{xxy23@mails.tsinghua.edu.cn}

\author[2]{ \sur{Peng Chen}}
\author[1]{\sur{Ben Qi}}
\author[1]{\sur{Hongru Zhao}}
\author*[1]{\sur{Jingang Liang}}\email{jingang@tsinghua.edu.cn}
\author[1]{\sur{Jiejuan Tong}}
\author[1]{\sur{Haitao Wang}}

\affil[1]{\orgdiv{Institute of Nuclear and New Energy Technology}, \orgname{Tsinghua University}, \orgaddress{ \city{Beijing}, \postcode{100084}, \country{China}}}       

\affil[2]{\orgdiv{Software Institute}, \orgname{Chinese Academy of Sciences}, \orgaddress{ \city{Beijing}, \postcode{100086},  \country{China}}}


\abstract{
Human reliability analysis (HRA) is crucial for evaluating and improving the safety of complex systems. Recent efforts have focused on estimating human error probability (HEP), but existing methods often rely heavily on expert knowledge, which can be subjective and time-consuming. Inspired by the success of large language models (LLMs) in natural language processing, this paper introduces a novel two-stage framework for knowledge-driven reliability analysis, integrating IDHEAS and LLMs (KRAIL). This innovative framework enables the semi-automated computation of base HEP values. Additionally, knowledge graphs are utilized as a form of retrieval-augmented generation (RAG) for enhancing the framework’s capability to retrieve and process relevant data efficiently. Experiments are systematically conducted and evaluated on authoritative datasets of human reliability. The experimental results of the proposed methodology demonstrate its superior performance on base HEP estimation under partial information for reliability assessment.}

\keywords{Human Reliability Analysis (HRA), Retrieval-Augmented Generation (RAG), Base Human Error Probability (BHEP),Large Language Models (LLMs) }



\maketitle

\section{Introduction}\label{Introduction}

Human reliability analysis (HRA) is a systematic methodology for evaluating the likelihood of human errors within complex systems, particularly in high-risk industries such as nuclear power, aviation, and healthcare. These fields require meticulous safety standards, where even minor human errors can lead to catastrophic outcomes \cite{xiao2024emergency}. The current HRA methods have evolved into a third generation, with the most prominent approaches being the IDHEAS-G\cite{chang2016general} and IDHEAS-ECA \cite{xing2020integrated} methods proposed by the U.S. Nuclear Regulatory Commission (NRC). These methods have contributed significantly to the field, with extensive and meaningful work. Both are based on IDHEAS-DATA \cite{xing2021draft} for building HRA models. Despite its importance, accurately estimating human error probability (HEP) remains a significant challenge. Compared to IDHEAS-DATA, these two methods are somewhat coarse, with a slight reduction in precision. Moreover, the process of solving the base HEP is manual and time-consuming, making it particularly challenging for newcomers. These challenges highlight the need for innovative approaches to HEP estimation that are both resource-efficient and accurate.

Large language models (LLMs) are advanced AI-driven systems trained on vast amounts of textual data to understand and generate human-like language. In recent years, LLMs have shown significant potential to improve data-driven decision-making across various domains, including high-risk industries \cite{cascella2023evaluating,dubovik2024advanced,xiao2024text}. By leveraging natural language processing capabilities, LLMs can process and analyze complex textual information, offering insights that support critical decision-making processes. For instance, Eigner et al. \cite{eigner2024determinants} explore the determinants of LLM-assisted decision-making. Chiang et al. \cite{chiang2024enhancing} investigate how LLM-powered Devil's Advocate can enhance AI-assisted group decision-making. Ma et al. \cite{ma2024towards} discuss the design and evaluation of LLM-empowered deliberative AI for human-AI deliberation in AI-assisted decision-making. However, there has been no research exploring the application of LLMs in supporting decision-making processes for deriving base HEP. Their potential use in HRA could provide powerful tools for analyzing and interpreting complex human performance data, thereby offering deeper insights into the dynamics of human error. Consequently, this paper seeks to leverage LLMs to semi-automate the generation of base HEP.

In this article, a novel LLM-based framework called knowledge-driven reliability analysis integrating IDHEAS and LLMs (KRAIL) is proposed for base HEP estimation. The contributions and novelties of this article are summarized as follows:
\begin{itemize}
\item  \textbf{An innovative LLM-based two-stage framework:} We propose a novel LLM-based two-stage for base HEP computation. The two stages are the multi-agent framework for task decomposition and the integration framework for base HEP calculation. 
\item \textbf{A highly efficient alternative to the conventional manual method for computing base HEP:} Traditional base HEP analysis methods are time-consuming and resource-intensive, resulting in low efficiency. The proposed KRAIL addresses this gap by enabling semi-automated computation of base HEP, significantly improving efficiency.
\item \textbf{Knowledge graph-enhanced base HEP technology:} We employ the knowledge graph to construct the framework. This approach efficiently facilitates the utilization of IDHEAS-DATA. It also enables the incorporation of domain expert knowledge into the KRAIL. Additionally, it plays a crucial role in the final error rate calculation process.
\end{itemize}

The remaining paper is organized as follows. Section 2 presents a literature review, covering human reliability analysis, research applications utilizing IDHEAS-DATA, large language models, and retrieval-augmented generation. Section 3 introduces the development of the proposed framework in detail. Section 4 demonstrates the experimental results and discusses the experimental performance. Finally, Section 5 concludes this paper.

\section{Literature Review}\label{Literature Review}

In this section, we introduced human reliability analysis, research applications utilizing integrated decision-making and human error analysis system database (IDHEAS-DATA), and the primary technologies used in this study: large language models and retrieval-augmented generation (RAG) techniques.

\subsection{Human Reliability Analysis}
\label{Human Reliability Analysis}

Human reliability analysis (HRA) is a multidisciplinary domain that evaluates the likelihood of human errors and their potential impact on complex systems, particularly in high-risk industries such as nuclear power, aviation, and healthcare \cite{di2013overview}. HRA methods are broadly categorized into first-generation, second-generation, and third-generation approaches. The first generation of HRA methods emerged between the 1960s and the mid-1980s \cite{cuschieri2010human}. They focus on static models of human error probability (HEP) estimation and include methods such as THERP (Technique for Human Error Rate Prediction) \cite{kirwan1996validation}, HCR (Human Cognitive Reliability) \cite{hannaman1985model}, ASEP (Accident Sequence Evaluation Program) \cite{swain1987accident}, etc.

The second generation of HRA methods emerged in the early 1990s. These methods focus on the dynamic cognitive processes involved in emergencies, such as detection, diagnosis, and decision-making, aiming to explore the mechanisms behind human errors \cite{groth2019hybrid}. They combine cognitive reliability assessment with action execution reliability. Representative models include the CREAM (Cognitive Reliability and Error Analysis Method) \cite{hollnagel1998cognitive}, ATHEANA (A Technique for Human Event Analysis) \cite{cooper1996technique}.

The third generation of HRA has partially shifted towards dynamic approaches, exemplified by methods such as integrated human event analysis system for event and condition assessment (IDHEAS-ECA) \cite{xing2020integrated}. The IDHEAS-ECA was introduced by the U.S. NRC in October 2022. This method provides a structured, step-by-step framework for analyzing human actions and their contextual factors. Human actions are modeled through five macrocognitive functions: detection, understanding, decision-making, action execution, and interteam coordination. The failure of human action is represented using a set of cognitive failure modes and performance-influencing factors. These components are systematically applied to calculate the human error probability (HEP). The method is supported by authoritative data, known as IDHEAS-DATA \cite{xing2021draft}, which ensures both its reliability and applicability.

\begin{figure}[h]
\centering
\includegraphics[width=1.0\textwidth]{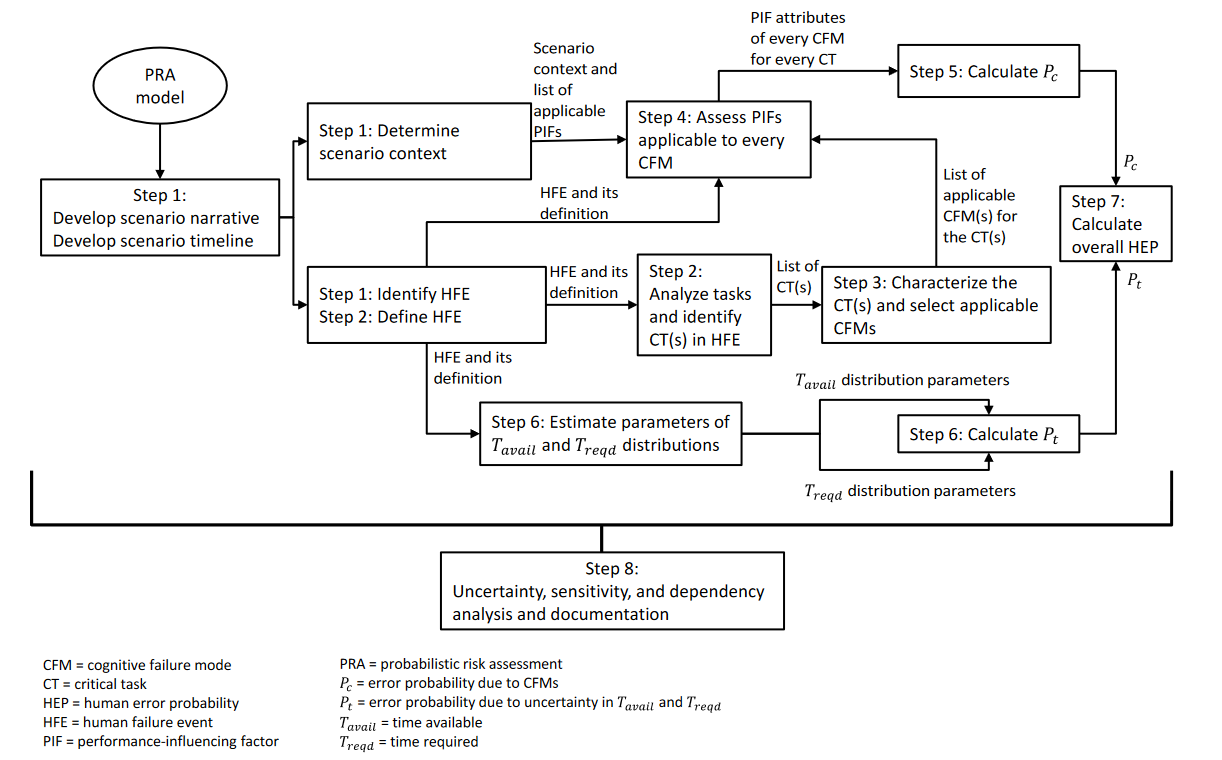}
\caption{IDHEAS-ECA HRA Process \cite{xing2020integrated} }\label{IDHEAS}
\end{figure}

The detailed IDHEAS-ECA process is illustrated in figure \ref{IDHEAS}. It consists of eight distinct steps. Although the IDHEAS-ECA \cite{xing2020integrated} method represents the most advanced approach among third-generation HRA methods, its granularity has changed significantly compared to IDHEAS-DATA \cite{xing2021draft}. For instance, the determination of the base HEP in IDHEAS-ECA requires only the assessment of performance influencing factors (PIFs) and cognitive failure modes (CFMs). This simplification introduces certain discrepancies compared to IDHEAS-DATA. IDHEAS-DATA includes five dimensions: PIF, CFM, task (and error measure), PIF measure, and other PIFs (and uncertainty). These categories are more detailed and precise, enabling a finer classification and analysis of the data. Additionally, in practice, most attributes, such as PIFs and CFMs, are determined through expert judgment. This reliance increases the subjectivity of the method, making it difficult to standardize and apply uniformly. Furthermore, the lack of a systematic approach to rapidly provide parameter recommendations poses additional challenges.

To address these limitations, this study proposes a method called KRAIL to quickly estimate attributes, thereby supporting experts in performing HRA more efficiently and consistently.

\subsection{Research Applications Utilizing IDHEAS-DATA}
The integrated human event analysis system for human reliability data (IDHEAS-DATA) provides a comprehensive dataset for understanding human error mechanisms in complex, high-risk environments. The data are organized into 27 tables, collectively known as the IDHEAS-DATA tables (IDTABLEs). Among these, IDTABLEs 1-3 specifically address base Human Error Probabilities (HEPs).IDTABLE-1 focuses on base HEPs for scenario familiarity. IDTABLE-2 presents base HEPs for information availability and reliability. IDTABLE-3 covers base HEPs for task complexity. The IDHEAS system is grounded in a solid foundation of literature \cite{levine2024identifying}, ensuring its reliability and relevance in the field of human reliability analysis.

Due to the relatively recent development of IDHEAS, there is limited related work. However, some scholars have referred to IDHEAS for human reliability data collection efforts. For instance, Markus Porthin et al. \cite{porthin2024task} proposed a task reliability index for operator performance and failure probability assessment in control room simulators. Additionally, other researchers have utilized the publicly available IDHEAS data to investigate third-generation dynamic HRA methods. An example is Jooyoung Park \cite{park2021dynamic}, who developed the model of human actions using event modeling risk assessment with linked diagrams (EMRALD). This dynamic simulation tool for PRA features a web-based graphical user interface for modeling and an open framework for easy coupling with physics codes. Furthermore, Park contributed to the development of the procedure-based investigation method of the EMRALD risk assessment—human reliability analysis (PRIMERA-HRA) method.

However, as evidenced by the IDHEAS-ECA tools, no auxiliary tools have been developed for base human error probability (HEP) analysis to date \cite{xing2020integrated}. Consequently, the method proposed in this study, KRAIL, aims to address this gap in the existing literature and practice.

\subsection{Large Language Models and Retrieval-Augmented Generation}

Recent advancements in large language models (LLMs), such as GPT and Claude, have demonstrated the ability to synthesize sophisticated understanding, such as translating between languages for which they have not been explicitly trained \cite{reid2024gemini}. Furthermore, LLMs can effectively learn from context, where new information—such as a grammar manual—enables the acquisition of new capabilities \cite{buehler2024accelerating}. As a result, LLMs have been widely applied to various tasks. For example, Xiao et al. \cite{xiao2024text} fine-tuned the nuclear large language model fault tree generator (NuLLM-FTG) to automate the construction of fault trees. Similarly, Li et al. \cite{li2024agent} used a multi-agent framework to simulate tasks in a hospital setting, including registration, consultation, medical examination, diagnosis, treatment recommendation, and convalescence. These examples illustrate the versatility and potential of LLMs in complex, specialized domains.

Retrieval-augmented generation (RAG) \cite{oettinger1990rag} is an advanced natural language processing (NLP) technique that combines two key components: a retrieval mechanism and a generative model. This hybrid architecture leverages external knowledge sources to enhance the generation of accurate and contextually relevant responses, making it particularly useful for tasks requiring domain-specific knowledge or up-to-date information. RAG integrates external knowledge sources, such as databases, domain-specific documents, or web-based repositories, into the generative process. During inference, relevant information is retrieved from these sources and provided as context to the LLM, enhancing its ability to generate accurate and contextually relevant responses. Many researchers have employed RAG to enhance the application of LLMs in specialized domains. For instance, Wang et al. \cite{wang2024biorag} introduced BioRAG, a RAG-LLM framework for biological question reasoning, while Li et al. \cite{li2024enhancing} explored the enhancement of LLM factual accuracy with RAG to mitigate hallucinations, specifically focusing on domain-specific queries in private knowledge bases. Inspired by these advancements, we aim to leverage RAG to enhance the analytical capabilities of LLMs in the context of human error probability analysis.

\section{Methodology}
This section provides a detailed explanation of the knowledge-driven reliability analysis integrating IDHEAS and LLMs (KRAIL) method. Specifically, the KRAIL method consists of two parts. The first part involves a multi-agent framework for task decomposition. The second part utilizes large language models (LLMs) and knowledge graphs to generate the attributes for determining the base HEP. This approach facilitates faster and more accurate analysis of the base HEP.

\subsection{Overview of the Proposed KRAIL}
\label{Overview of KRAIL}
As discussed in Section \ref{Human Reliability Analysis}, IDHEAS-DATA offers a more precise method for base HEP calculation compared to IDHEAS-ECA. Therefore, the proposed method, KRAIL, is based on the method outlined in IDHEAS-DATA \cite{xing2018use}. The application and data collection process for IDHEAS-DATA consists of five essential steps. First, the data source is analyzed to understand the context and identify human error data suitable for generalization. Next, task analysis is performed to determine the relevant cognitive failure modes (CFMs). The context is then mapped to relevant performance influencing factor (PIF) attributes, while additional PIF attributes present in the study are also identified. Finally, uncertainties are assessed, and the reported human error data are systematically documented in the IDTABLE.

\begin{figure}[h]
\centering
\includegraphics[width=1.0\textwidth]{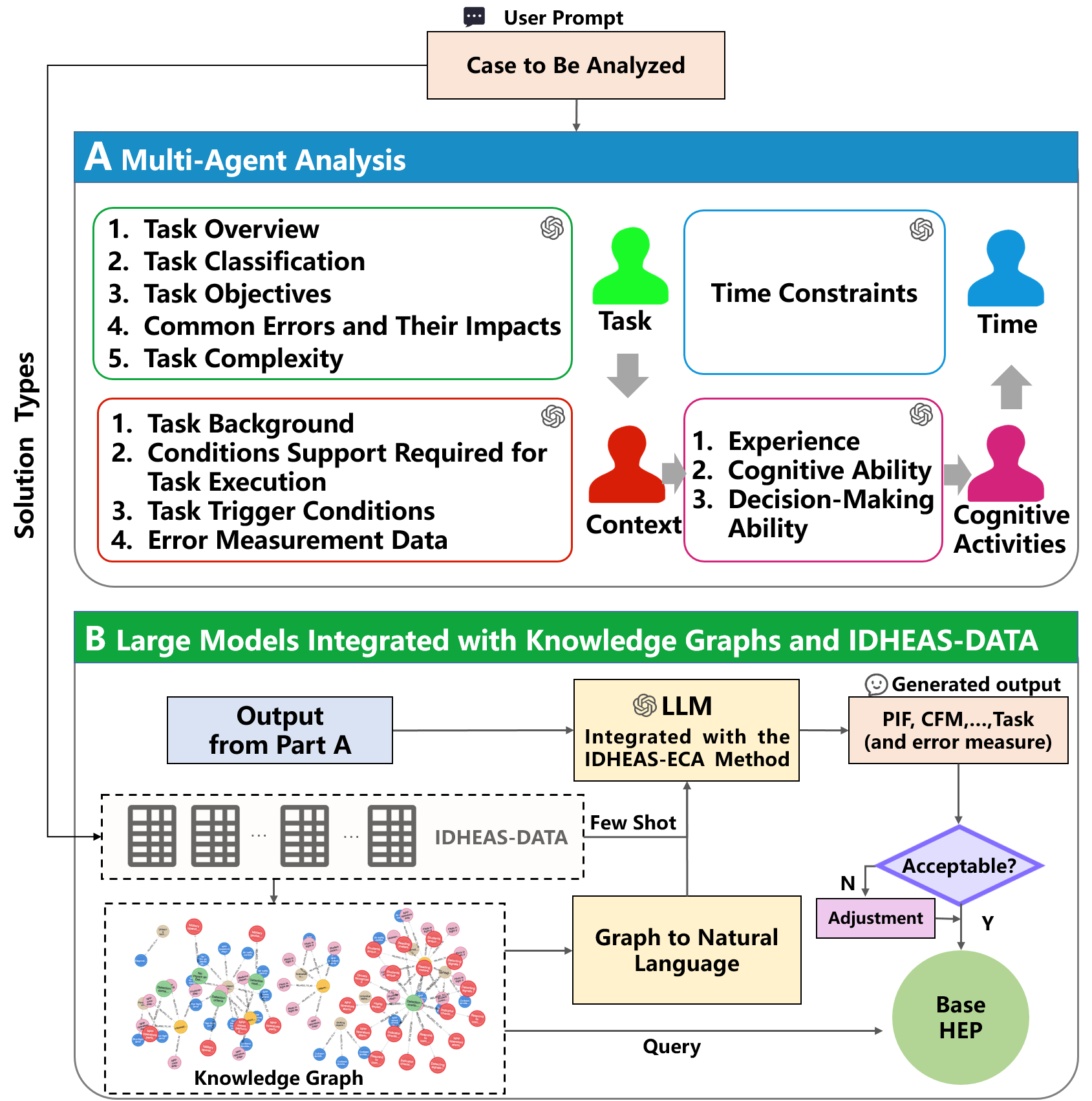}
\caption{KRAIL: A Knowledge-Driven Framework Integrating IDHEAS and Large Language Models for Base Human Reliability Analysis}\label{flow}
\end{figure}

Leveraging the capabilities of large language models (LLMs), the analysis process is both accelerated and simplified. The proposed approach organizes the workflow into two main steps. First, the user inputs the case information for analysis. This data is then processed through a task decomposition framework using a multi-agent approach, as illustrated in Figure \ref{flow}, part A. Next, the selected base human error probability (HEP) solution type is entered into part B. The second step involves the extraction of IDHEAS-DATA attributes, utilizing the logical relationships between knowledge graph nodes, as shown in Figure \ref{flow}, part B. By integrating LLMs, knowledge graph networks, and expert knowledge, the final base HEP is determined. The following sections offer a detailed explanation of each component within this framework.

\subsection{Multi-Agent Framework for Task Decomposition}
\label{Multi-Agent Framework for Task Decomposition}

This section primarily introduces part A of KRAIL framework, which focuses on the multi-agent analysis process. Four LLM-driven agents are deployed to perform distinct tasks: task analysis, context analysis, cognitive activities analysis, and time constraints analysis. After the case information is input, these agents are designed to further decompose and analyze the data.

\textbf{Agent 1. Task Analysis.} The primary task is to analyze the data source to identify tasks associated with reported human error information. It encompasses several key components. First, a task overview is conducted to outline the general process of the task. Second, a task classification is performed, categorizing tasks based on their functions or types to facilitate a clearer understanding of their nature and context. Third, the objectives of the task are analyzed to define the specific goals associated with each task. Fourth, an examination of typical error types and their impacts is carried out to identify common errors within the tasks and assess their potential consequences. Finally, the task complexity level is determined, providing insight into the degree of difficulty involved, ranging from simple to highly complex tasks.

\textbf{Agent 2. Context Analysis.} The core task is to analyze the data source and the context associated with task execution. Specifically, this involves four key tasks: first, identifying the background conditions under which the task occurs; second, analyzing the support required for task execution; third, clarifying the initial conditions and requirements for task initiation; and finally, examining the error measurement data to assess task-related performance and outcomes.

\textbf{Agent 3. Cognitive Activities Analysis.} The primary task is to analyze the data source and characterize the tasks to identify the specific cognitive activities involved. This includes examining the cognitive demands required for task execution and understanding the mental processes underlying task performance.

\textbf{Agent 4. Time Constraints Analysis.} The primary task is to analyze the data source and evaluate the time constraints associated with task execution. This involves identifying the temporal limitations, deadlines, or time-sensitive conditions that impact task performance and outcomes.

\subsection{Integration Framework for Base HEP Calculation}

This section primarily focuses on part B of the KRAIL framework. As illustrated in Figure \ref{flow}, it leverages knowledge graphs, large language models (LLMs), IDHEAS-DATA, and expert knowledge.

Initially, the user selects the analysis type for the base human error probability (HEP), choosing from options such as scenario familiarity, information availability and reliability, or task complexity. Based on this selection, Part B of the IDHEAS-DATA is dynamically adjusted to align with the user's choice. Consequently, the resulting knowledge graph also changes. To enhance this process, we generate a few-shot learning content using the IDHEAS-DATA. Few-shots learning is a common technique to augment the capabilities of LLMs by including examples in the input prompt \cite{ahmed2022few}. This helps the model better understand the problem. Common variations of few-shot learning include 0-shot, 1-shot, 3-shot, and 5-shot.

Next, the output from part A, the few-shot content, and the natural language conversion of the knowledge graph are all fed into the LLM. This process results in the desired attributes: PIF, CFM, task (and error measure), PIF measure and other PIFs (and uncertainty). The output attributes are then presented for expert review. If the expert accepts the result, the relationships between the nodes in the knowledge graph are utilized to search for the final base HEP. If the expert does not accept the result, manual adjustments will be made, after which the relationships in the knowledge graph are again used to search for the final base HEP.

\section{Experimental Results}
This section presents the experimental results. First, a dataset was created based on the IDHEAS-DATA. Subsequently, experimental setups and preparations were conducted. The performance was compared across different shot configurations, followed by a time comparative analysis between KRAIL and manual methods. An ablation experiment was also performed, along with a case study.

\subsection{Data Setup}
Due to the scarcity of HRA data and the lack of publicly available case data in the IDHEAS-DATA, we referred to the IDHEAS-DATA to prepare a dataset. Specifically, we use LLMs and expert knowledge to generate a high-quality HRA case dataset. To illustrate the data collection process, a sample was conducted. Appendix Table A1-2 in the IDEHAS-DATA documentation presents table examples, as shown in Table \ref{Table_A1_2}.

\begin{table}[h]
\caption{Appendix Table A1-2 in the IDEHAS-DATA Documentation (Samples) \cite{xing2021draft}}\label{Table_A1_2}%
\begin{tabular}{@{}p{0.6cm}p{0.6cm}p{1cm}p{3.6cm}p{2cm}p{1.6cm}p{0.6cm}@{}}
\toprule
PIF& CFM &Error rates & Task (and error measure)& PIF Measure & Other PIFs (and Uncertainty)&
REF\\
\midrule
SF3.3 & DM&  0.5&  Medicine dispensing Lack of plans, 
policies and procedures to address the situation& 
Inadequate time, Training, procedure &
& ref \cite{cohen2012risk}\\
SF4  &D & 0.2 & Railroad operators start new workshift (fail to check hardware unless specified)  &New workshift, task not specified so no mental model for checking & (Other PIF may exist)  & ref \cite{mcguirl2006supporting} \\
SF0 & U & 1.6E-3 & Situation assessment in EOP (HEP of Critical Data Dismissed/Discounted)  &Inappropriate Bias not formed, No Confirmatory Information & (Expert judgment)  &ref \cite{xing2017integrated}\\
SF4 & U & 2.5E-1  & Situation assessment in EOP (HEP of Critical Data Dismissed/Discounted)  &Inappropriate Bias formed, No Confirmatory Information  &(Expert judgment)  &ref \cite{xing2017integrated} \\
SF0 & U  &3.5E-4 & Critical Data Collection (Premature Termination of Critical Data Collection) & Expectations or Biases not formed & (Expert judgment) & ref \cite{xing2017integrated}\\
SF4 & D / U & 8.2E-3 & Critical Data Collection (Premature Termination of Critical Data Collection)  &Expectations or Biases formed  & The failure mode could be either D or U. (Expert judgment)  & ref \cite{xing2017integrated} \\
\botrule
\end{tabular}
\end{table}

Next, we referred to the contents of the references in Table \ref{Table_A1_2} and identified the original literature sources (ref \cite{cohen2012risk, mcguirl2006supporting,xing2017integrated}), for analysis. We use ref \cite{cohen2012risk} as an example to demonstrate our analytical process. In ref \cite{cohen2012risk}, the relevant data identified are presented in Table \ref{Case122}. Subsequently, the preliminary data were input into the LLM, which identified cases 4, 5, and 6 as corresponding to the data. This judgment was validated by domain experts. If the analysis provided by the LLM was incorrect, experts utilized their domain-specific knowledge to make corrections. As a result, parameters such as PIF, CFM, error rates, task (and Error Measure), PIF measure, and other PIFs (and Uncertainty) were assigned the values specified in Table \ref{Table_A1_2}.

\begin{table}[h]
\caption{Descriptions of Human Error Data from Cohen (2012)\cite{cohen2012risk}}\label{Case122}%
\begin{tabular}{@{}lp{12cm}@{}}
\toprule
Index & Task \\
\midrule
Case 1 & Wrong drug disDispensing errors: (1) Wrong drug selected when pensed manually filling a warfarin prescription. (2) Wrong drug selected when filling an automated dispensing cabinet with warfarin.\\
Case 2 & Wrong dose/strength Prescribing error: Wrong dose/strength tablets of warfarin dispensed prescribed.
Wrong dose/strength Dispensing errors: (1) Wrong warfarin dose/ of warfarin dispensed strength selected when manually filling a warfarin prescription. (2) Wrong warfarin dose/strength selected when filling an automated dispensing cabinet with warfarin.\\
Case 3 &Wrong dose/strength Dispensing error: Wrong dose/strength selected or of warfarin dispensed entered during data entry of a warfarin prescription.\\
Case 4 &Warfarin dispensed with wrong directions for use Prescribing error: Warfarin prescription included directions to take the drug more often than daily.\\
Case 5 &Warfarin dispensed Dispensing error: Warfarin prescription entered to the wrong patient into the wrong patient’s drug profile.\\
Case 6 &Warfarin dispensed Dispensing errors: (1) Warfarin vial placed in a bag to the wrong patient containing another patient’s medications. (2) Wrong patient’s medication(s) selected from the will-call area at the point of sale.
 \\
\botrule
\end{tabular}
\end{table}

It is worth noting that, due to resource limitations, a comprehensive review of all relevant literature was not feasible. Instead, we focused on ref \cite{preischl2013human,preischl2016human,aalipour2016human,cohen2012risk,lee2002collision}, extracting and organizing data from these sources. From the curated dataset, 59 entries were collected for PIF attributes and base HEPs for scenario familiarity, 11 entries for PIF attributes and vase HEPs for information availability and reliability, and 26 entries for PIF attributes and base HEPs for task complexity.

\subsection{Experimental Setup}

Next, the following provides a detailed description of the experimental setup. In terms of LLM selection, we have chosen Claude 3.5 Sonnet. Claude 3.5 has outperformed models such as GPT-4, Gemini 1.5, and Llama-400B in most benchmarks. In code-related benchmarks, Claude 3.5 achieved a remarkable 92\% score in zero-shot tasks. It has also earned top scores in areas including visual mathematical reasoning, scientific table processing, chart-based question answering, and document-based question answering \cite{kurokawa2024diagnostic}.

Additionally, the tables from IDHEAS-DATA for base HEP, including Appendix A1 (PIF attributes and base HEPs for scenario familiarity), Appendix A2 (PIF attributes and base HEPs for information availability and reliability), and Appendix A3 (PIF attributes and base HEPs for task complexity), were organized and structured. A knowledge graph was constructed in Neo4j \cite{miller2013graph} to analyze the base HEP. Neo4j is a graph-oriented database type, a highly efficient data storage structure. The detailed data from the IDHEAS-DATA tables are thoroughly explained in ref \cite{xing2021draft} and will not be repeated in this paper.

Additionally, to ensure accuracy, our study focuses on auxiliary operations and employs the top-5 metric. The top-5 output refers to the top five ranked results or data points \cite{campochiaro2009metrics}. If the correct answer appears within these top five results, it is considered correct.

\subsection{Performance Across Different Shot Configurations}

To demonstrate the confidence intervals of our data, the bootstrap distribution method was employed. This was applied to assess the LLM performance across different shot configurations under varying scenario conditions. The bootstrapping technique allowed for the estimation of the distribution of performance metrics, providing a more robust understanding of the model's reliability across different settings \cite{mooney1993bootstrapping}.

\begin{figure}[H]
\centering
\includegraphics[width=1.0\textwidth]{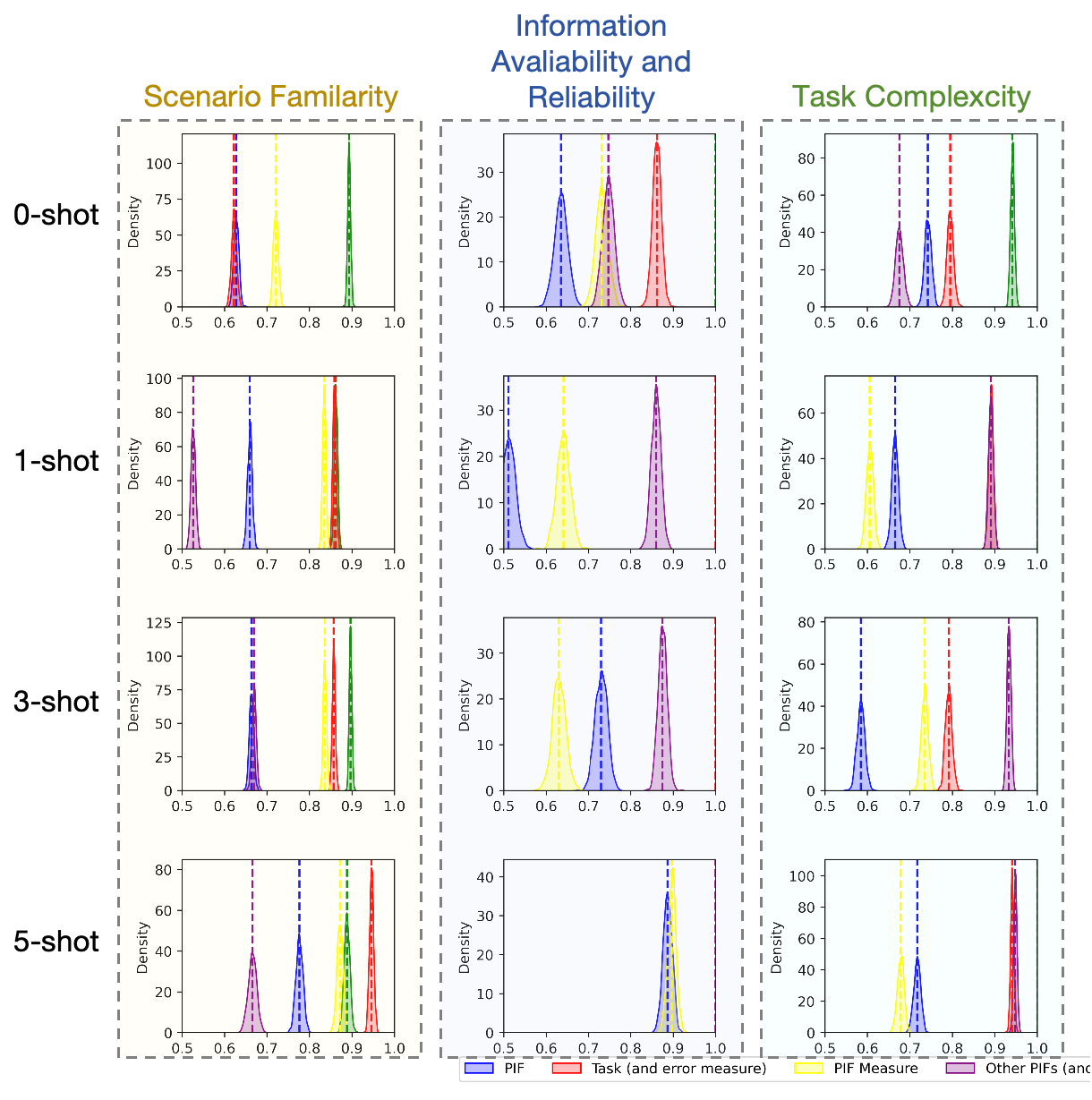}
\caption{Performance Across Different Shot Configurations for Varying Scenario Conditions. }\label{fews}
\end{figure}

The yellow section in figure \ref{fews} represents the accuracy distribution for scenario familiarity, while the blue section depicts the accuracy distribution for information availability and reliability. The green section shows the accuracy distribution for task complexity. Each row corresponds to the LLM performance across 0-shot, 1-shot, 3-shot, and 5-shot settings. From figure \ref{fews}, it can be observed that the LLM achieves the highest performance across all three dimensions in the 5-shot setting. This indicates that the LLM's accuracy improves with an increasing number of examples, reaching its optimal performance at 5-shot.

Next, we computed the mean $\pm$ standard deviation (std) for five dimensions: PIF, CFM, task (and error measure), PIF measure, and other PIFs (and uncertainty). The tables \ref{table1,table2,table3} provide a detailed examination of the LLM performance across 0-shot, 1-shot, 3-shot, and 5-shot under scenario familiarity, information availability and reliability, and task complexity. 

As for scenario familiarity, the general trend shows improvements in LLM performance with increasing shots across most metrics. Specifically, PIF increases from $0.628 \pm 0.119$ (0-shot) to $0.777 \pm 0.171$ (5-shot), indicating that more examples help the model better understand the scenario's context. Performance significantly improves from $0.622 \pm 0.115$ (0-shot) to $0.946 \pm 0.096$ (5-shot), reflecting the positive impact of more examples on task-specific performance. CFM remains relatively stable, slightly decreasing from $0.893 \pm 0.069$ (0-shot) to $0.888 \pm 0.132$ (5-shot), suggesting that familiarity with the scenario is robust across shot configurations. Other PIFs improve consistently from $0.374 \pm 0.112$ (0-shot) to $0.666 \pm 0.204$ (5-shot), showing that more examples allow the model to generalize better to various PIF-related tasks. As expected, the model benefits from more shots. However, the increased variability in higher shot configurations (e.g., PIF and Other PIFs) suggests that the model might experience some instability or complexity as it processes more examples.

\begin{table}[h]
\caption{Mean$\pm$Standard Deviation of LLM Performance Across Different Shot Configurations for Scenario Familiarity}\label{table1}
\begin{tabular*}{\textwidth}{@{\extracolsep\fill}p{1cm}p{2cm}p{2cm}p{2cm}p{2cm}p{2cm}}
\toprule%
Project & PIF & CFM	& Task & PIF Measure & Other PIFs \\
\midrule
 0-shot & $0.628\pm0.119$  & $0.893\pm0.069$ &$0.622\pm0.115$ &$0.720\pm0.112$ &$0.374\pm0.112$ \\
1-shot& $0.659\pm0.106$  & $0.862\pm0.079$  & $0.858\pm0.078$ &$0.835\pm0.076$  & $0.526\pm0.103$   \\
3-shot& $0.663\pm0.103$  & $0.896\pm0.064$ & $0.857\pm0.072$ & $0.836\pm0.075$ & $0.669\pm0.103$  \\
5-shot& $0.777\pm 0.171$ & $0.888\pm0.132$ & $0.946\pm0.096$ & $0.872\pm0.136$ & $0.666\pm0.204$   \\
\botrule
\end{tabular*}
\end{table}

Regarding information availability and reliability in table \ref{table2}, CFM remains constant at a perfect score of 1.000 across all shot configurations, reflecting the model’s consistent handling of information availability and reliability. PIF initially decreases from $0.635 \pm 0.317$ (0-shot) to $0.511 \pm 0.320$ (1-shot), before improving to $0.886 \pm 0.209$ (5-shot), indicating a marked increase in model performance as more examples become available. Performance increases from $0.861 \pm 0.223$ (0-shot) to $1.000 \pm 0.000$ (1-shot), with subsequent stabilization at $1.000$ in higher shot configurations (3-shot and 5-shot), suggesting that a small number of examples suffices for peak performance on tasks related to information availability and reliability. PIF Measure shows a moderate increase, peaking at $0.869 \pm 0.068$ (5-shot), while Other PIFs demonstrate a steady increase from $0.747 \pm 0.276$ (0-shot) to $1.000 \pm 0.000$ (5-shot), indicating that more shots allow the model to handle the complexities of various PIFs more effectively. It can be concluded that the model’s performance in handling information availability and reliability improves significantly with the introduction of more examples, especially after the 1-shot configuration. While CFM remains perfect throughout, other metrics show substantial improvements, particularly in other PIFs and performance, where a minimal number of examples is sufficient to reach peak performance.

\begin{table}[h]
\caption{Mean$\pm$Standard Deviation of LLM Performance Across Different Shot Configurations for Information Availability and Reliability}\label{table2}
\begin{tabular*}{\textwidth}{@{\extracolsep\fill}p{1cm}p{2cm}p{2cm}p{2cm}p{2cm}p{2cm}}
\toprule%
Project & PIF & CFM	& Task & PIF Measure & Other PIFs  \\
\midrule
 0-shot & $0.635\pm0.317$  & $1.000\pm0.000$ &$0.861\pm0.223$ &$0.732\pm0.286$ &$0.747\pm0.276$ \\
1-shot& $0.511\pm 0.320$  & $1.000\pm0.000$  & $1.000\pm0.000$ &$0.640\pm0.300$  & $0.860\pm0.224$   \\
3-shot& $0.731\pm0.288$  & $1.000\pm0.000$  & $1.000\pm0.000$ & $0.630\pm0.317$ & $0.875\pm0.216$  \\
5-shot& $ 0.886\pm0.209$ &  $1.000\pm0.000$  & $1.000\pm0.000$ & $ 0.869\pm0.068$ & $1.000\pm0.000$  \\
\botrule
\end{tabular*}
\end{table}

In terms of task complexity in table \ref{table3}, PIF decreases slightly from $0.742 \pm 0.161$ (0-shot) to $0.584 \pm 0.190$ (3-shot) before rising to $0.720 \pm 0.157$ (5-shot), indicating that task complexity has a more variable impact on performance compared to other scenarios. CFM improves from $0.941 \pm 0.091$ (0-shot) to a perfect score of 1.000 across all higher shot configurations, showing that the model adapts well to complex tasks once enough examples are provided. Performance increases significantly from $0.794 \pm 0.149$ (0-shot) to $0.943 \pm 0.079$ (5-shot), highlighting the importance of increasing the number of examples to handle more complex tasks effectively. PIF Measure shows steady but moderate improvement across all configurations, reaching $0.677 \pm 0.163$ (5-shot). Similarly, Other PIFs improve from $0.675 \pm 0.175$ (0-shot) to $0.943 \pm 0.079$ (5-shot), showing consistent improvements with more examples. The model’s ability to handle task complexity improves as more examples are provided, with substantial gains in performance and CFM stability across configurations. Despite initial variability in PIF, the model becomes more adept at managing complex tasks with increasing shots, suggesting that higher shot configurations provide a more robust understanding of task complexity.

\begin{table}[h]
\caption{Mean$\pm$Standard Deviation of LLM Performance Across Different Shot Configurations for Task Complexity}\label{table3}
\begin{tabular*}{\textwidth}{@{\extracolsep\fill}p{1cm}p{2cm}p{2cm}p{2cm}p{2cm}p{2cm}}
\toprule%
Project & PIF & CFM	& Task& PIF Measure & Other PIFs  \\
\midrule
 0-shot & $0.742\pm0.161$  & $0.941\pm0.091$ &$0.794\pm0.149$ &$0.394\pm0.195$ &$0.675\pm0.175$ \\
1-shot& $0.665\pm0.153$  & $1.000\pm0.000$  & $0.891\pm0.107$ &$0.605\pm0.163$  & $0.890\pm0.108$   \\
3-shot& $0.584\pm0.190$  & $1.000\pm0.000$& $0.791\pm0.162$ & $0.735\pm0.159$ & $0.933\pm0.094$  \\
5-shot& $0.720\pm 0.157$ & $1.000\pm0.000$ & $0.943\pm0.079$ & $0.677\pm0.163$ & $0.943\pm0.079$   \\
\botrule
\end{tabular*}
\end{table}

Across all three scenarios, the LLM's performance generally improves as the number of shots increases. However, the nature of the improvements varies across different tasks. In Scenario Familiarity, the model shows clear benefits in task-specific performance, with stable CFM values.
In Information Availability and Reliability, the model quickly adapts with few examples, achieving peak performance by the 1-shot configuration.
In task complexity, more examples help the model navigate complex tasks more effectively, with marked improvements across most metrics as shot configurations increase. These findings suggest that while more examples tend to improve LLM performance, the exact benefits depend on the task and scenario type. The model’s capacity for generalization becomes more apparent as the number of examples increases. The variability observed in some metrics at higher shot configurations underscores the need for further fine-tuning and optimization to ensure consistent performance across all domains.

\subsection{Time Comparative Analysis: KRAIL vs. Manual}

\begin{figure}[H]
\centering
\includegraphics[width=0.85\textwidth]{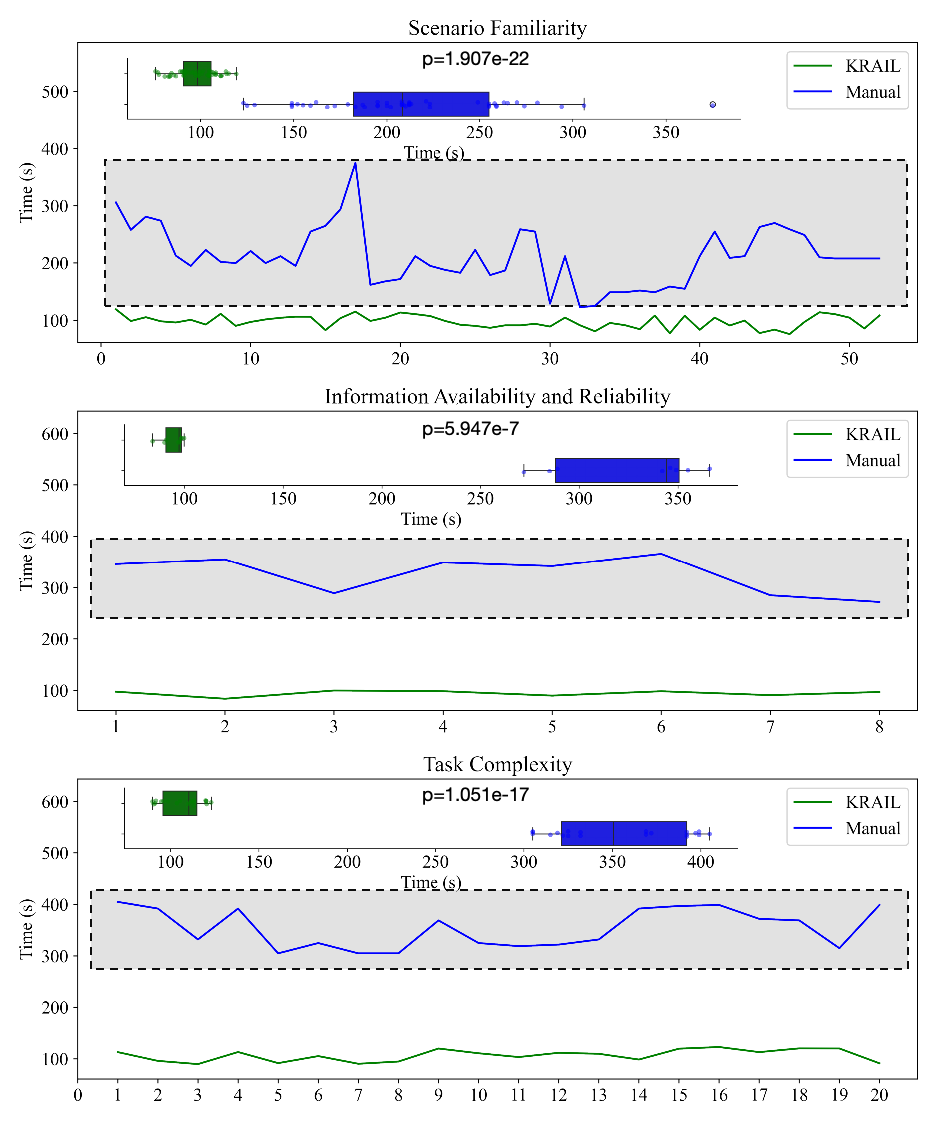}
\caption{Comparison of Time Spent Using Manual Methods vs. KRAIL Framework Across Different Scenarios}\label{time_compare}
\end{figure}

To demonstrate the efficiency and time-saving advantages of our algorithm, we compared the time required for a PhD student to classify input tasks manually with the time required by our KRAIL. The comparison covered three different tasks: scenario familiarity, information availability and reliability, and task complexity. The results are presented in Figure \ref{time_compare}. 

The upper part of each subplot shows boxplots of the time consumed using manual methods and the KRAIL system. Statistical significance (p-values) was calculated using a t-test to assess the distribution of the data. The p-values for the three tasks were $1.907e-22$, $5.947e-7$, and $1.051e-17$, respectively. These values are all below the standard significance threshold of 0.05, indicating that the observed differences are unlikely to have occurred by random chance \cite{kim2015t}. Therefore, the results demonstrate statistically significant differences, affirming the effectiveness and efficiency of KRAIL in reducing task classification time.

Additionally, it can be observed from the figure \ref{time_compare} that, for different task types, significant differences arise in human performance due to varying PIF attributes. However, for KRAIL, as the framework is automatically generated, the impact of task type on time consumption is minimal, with the mean time remaining around 100 seconds for all tasks. Relative to scenario familiarity, information availability and reliability and task complexity are more complex, leading to longer processing times. These tasks require more time compared to Scenario Familiarity, reflecting their higher complexity.
% 5-shot

\subsection{Ablation Experiment}

\begin{figure}[H]
\centering
\includegraphics[width=1.0\textwidth]{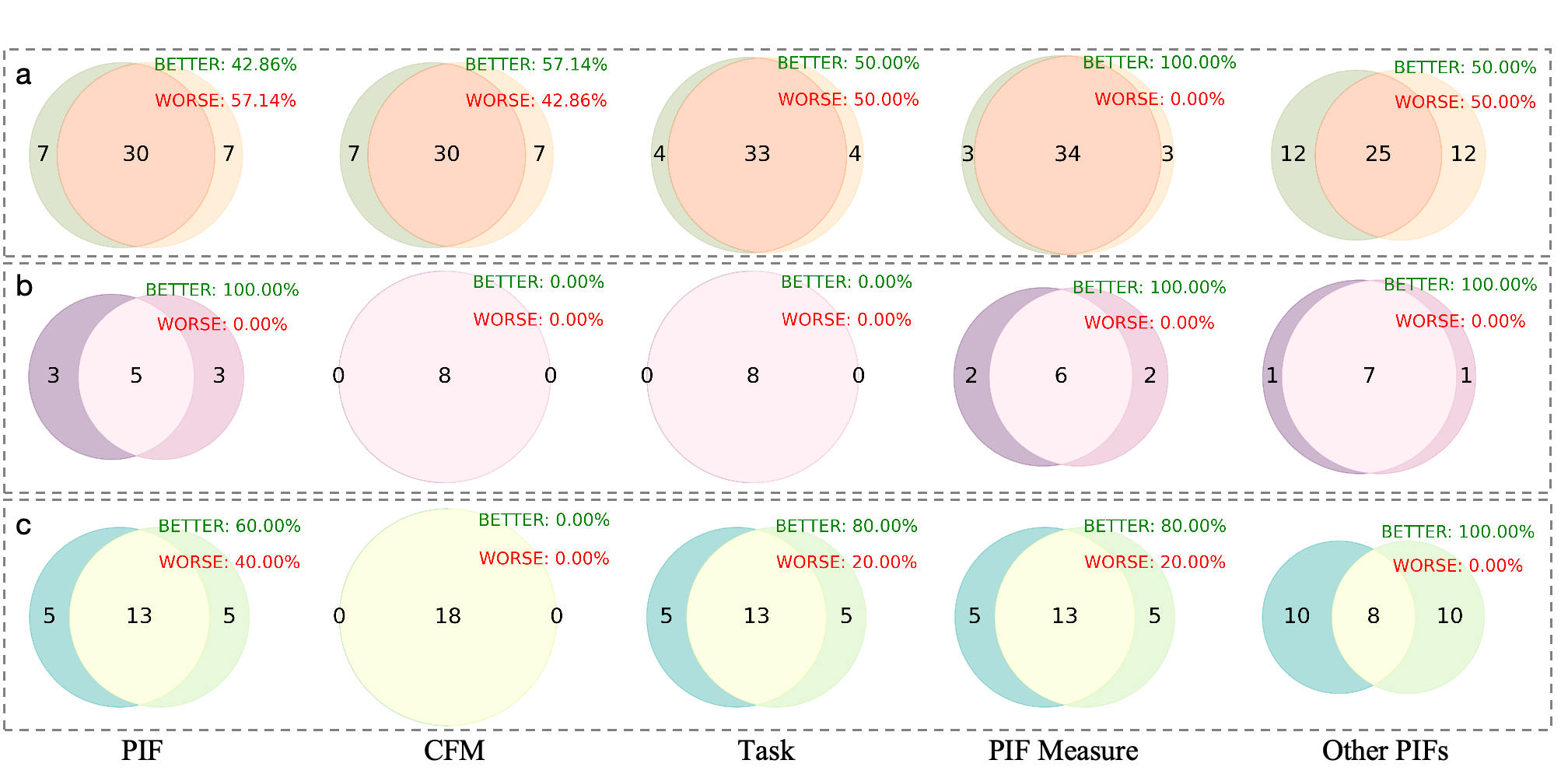}
\caption{Venn Diagram of Ablation Study: Impact of Removing Multi-Agent on Accuracy. The intersection area represents the results that remained unchanged after removing the multi-agent component. The proportions labeled BETTER and WORSE reflect the accuracy changes in different elements after the removal of the multi-agent framework. Specifically, if, for a given case, the judgment for PIF changed from "Yes" to "No" after removing the multi-agent component, that case was classified as a bad case. The WORSE value in the figure represents the ratio of bad cases to the total number of distinct elements that showed a change in accuracy. The higher the BETTER value, the more effective our multi-agent framework is.}\label{venns}
\end{figure}

Next, we conducted an ablation experiment to validate the effectiveness of the multi-agent framework described in Section \ref{Multi-Agent Analysis}. The results are presented in Figure \ref{venns}. 

Regarding scenario familiarity in Figure \ref{venns} (a), the introduction of multi-agent systems leads to a 42.86\% improvement and a 57.14\% deterioration in the PIF. In terms of CFM, 57.14\% of cases show improvement, while 42.86\% exhibit worse performance. Performance is evenly split, with 50\% showing improvement and 50\% showing deterioration. Regarding the PIF measure, 100\% of cases show improvement, with no deterioration, while other PIFs exhibit a 50\% improvement rate, with 50\% showing worse performance.

In terms of information availability and reliability in Figure \ref{venns} (b), the addition of multi-agent systems results in a 100\% improvement and no deterioration for PIF. For both CFM and task, no change in accuracy is observed, with performance remaining consistent before and after the integration of multi-agent systems. For the PIF Measure, again, there is 100\% improvement with no deterioration. Similarly, Other PIFs show a 100\% improvement rate, with no worsening observed.

For the task complexity PIF factor in Figure \ref{venns} (c), the introduction of multi-agent systems results in 60\% better performance and 40\% worse. No change is observed for CFM. For Task, 80\% of cases show improvement with 20\% showing deterioration, though overall accuracy remains unchanged. The PIF Measure shows 80\% better performance and 20\% worse. Lastly, Other PIFs again exhibit 100\% better performance, with no deterioration.

Overall, the differences across results are significant due to the varying references for determining PIF types based on different options. For the scenario familiarity condition, the multi-agent framework demonstrates a notable improvement in the PIF measure but shows limited enhancement for other factors. In contrast, for information availability and reliability, the designed multi-agent framework provides significant improvements in PIF, PIF measure, and other PIFs. However, its impact on CFM and performance is not evident. For task complexity, the multi-agent framework achieves notable improvements in PIF measure and other PIFs, highlighting its effectiveness in addressing the complexities associated with these dimensions.

\subsection{Case Study}
\begin{figure}[H]
\centering
\includegraphics[width=0.9\textwidth]{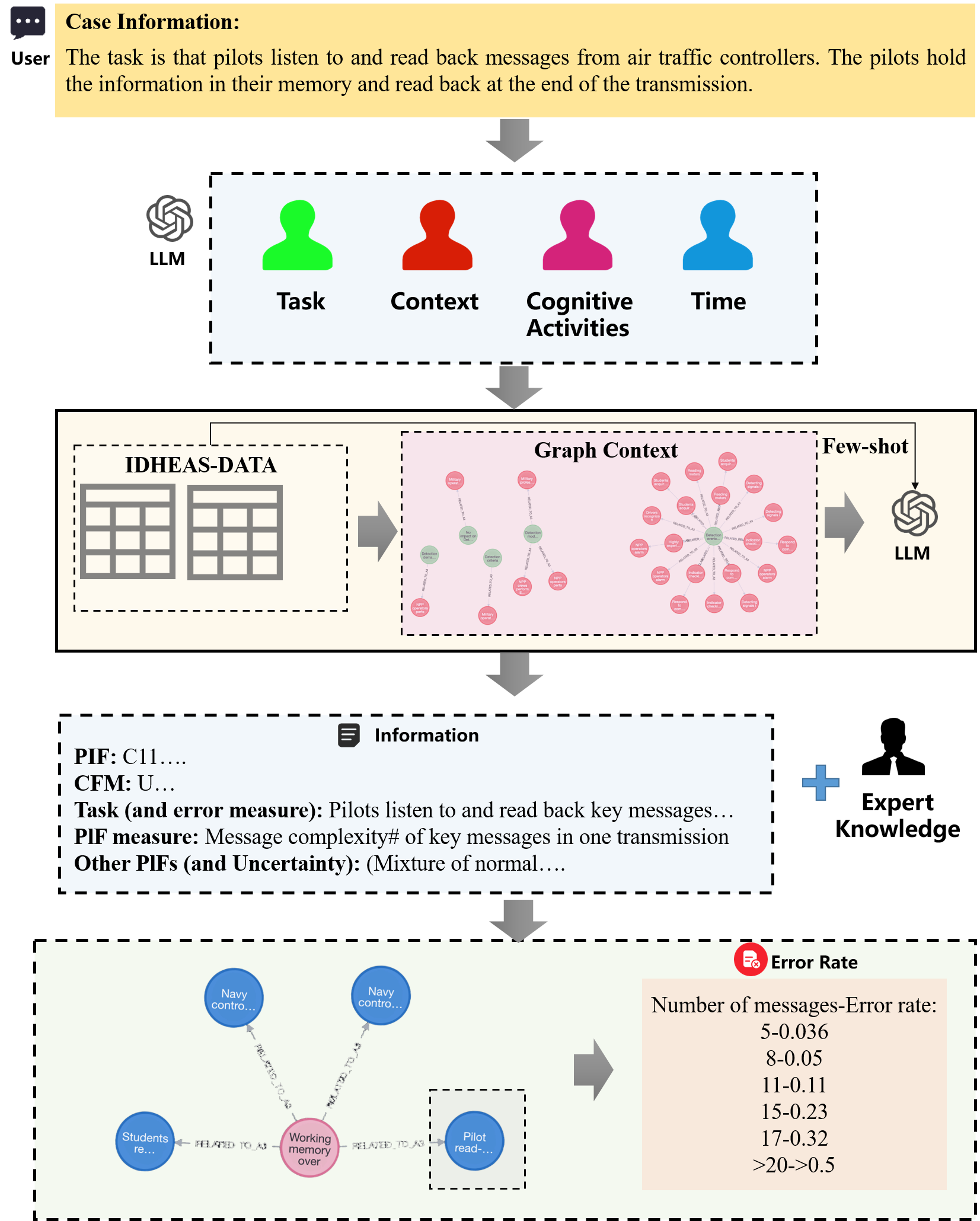}
\caption{Case Study Illustration of the KRAIL Framework. The yellow section in this figure represents the case information, which is then processed within our multi-agent framework. This is followed by integration with the graph context to produce the values of additional parameters. Finally, the error rate is determined. }\label{case_study}
\end{figure}

Lastly, to demonstrate the effectiveness of KRAIL. A case study is conducted. Specifically, a pilot communication task is selected as the test case. The provided case information is shown in Figure \ref{case_study}. The case information is then processed within the KRAIL multi-agent framework for detailed analysis. Subsequently, the graph context derived from IDHEAS-DATA is converted into natural language. This natural language input is fed into the LLM, which outputs additional parameter information for error rate output, as indicated by the green section in the figure. Finally, the graph is integrated to produce the final error rate. Due to space limitations, the outputs of task analysis, context analysis, cognitive activities analysis, and time constraints within the multi-agent framework are provided in Appendix \ref{secA1}.

\begin{figure}[H]
\centering
\includegraphics[width=1.0\textwidth]{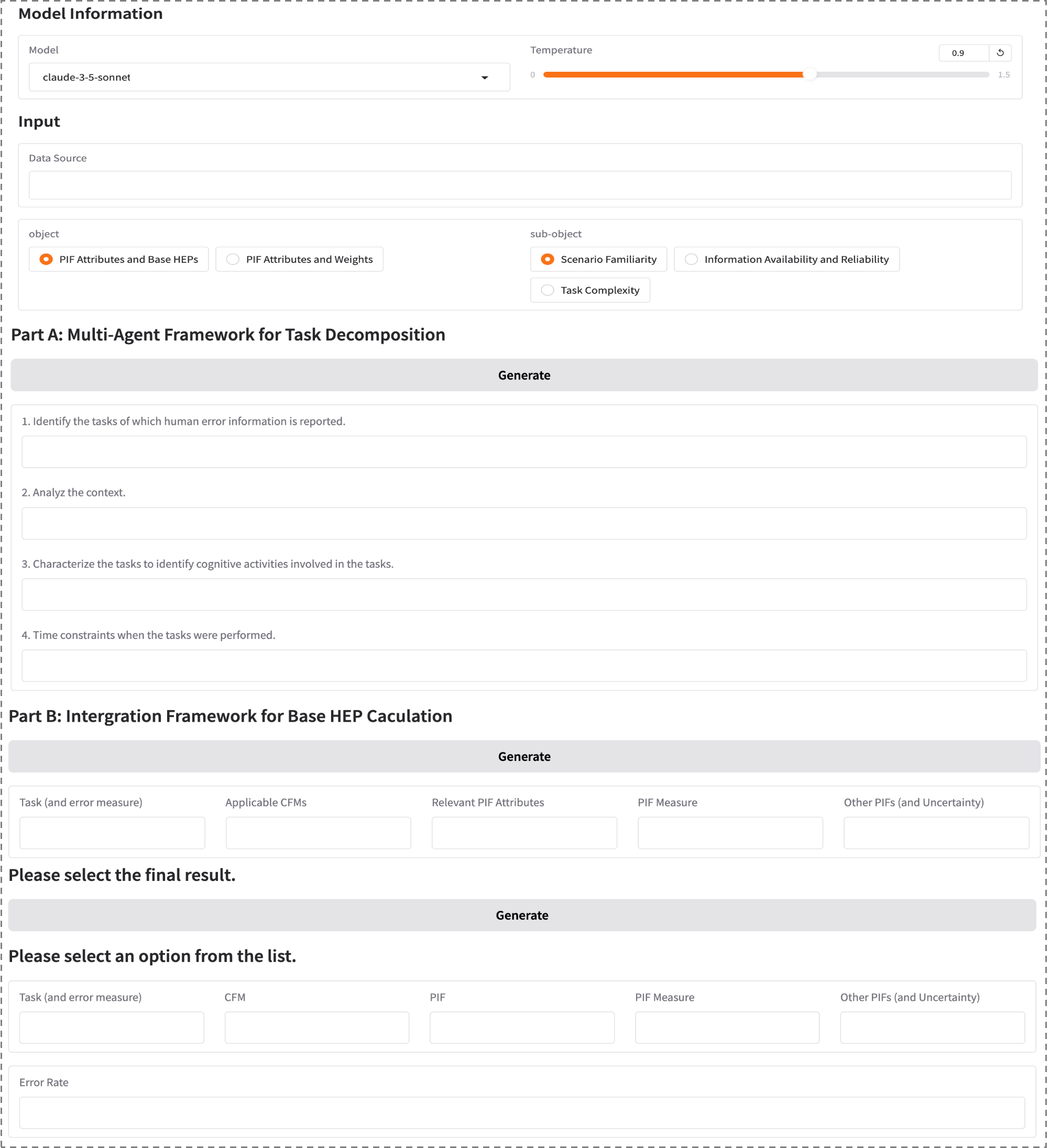}
\caption{KRAIL Framework Web Interface: User Interaction Dashboard. }\label{demo}
\end{figure}

To further extend the applicability of the KRAIL method, a web-based interface was developed using Gradio. This implementation provides an intuitive and user-friendly platform for end-user interaction. The specific interface is shown in Figure \ref{demo}. Users can first input information in the "Data Source" field, then select the solution type for base HEPs. Next, by clicking the "Generate" button, the backend triggers part A: the multi-agent framework for task decomposition. Afterward, the user clicks the second "Generate" button, initiating the algorithm for part B. This step generates attribute values such as task (and error measure), CFMs, PIF, and others. Finally, the expert can review and modify the generated attributes as needed. Once the adjustments are made, the expert clicks the last "Generate" button, which invokes the Neo4j knowledge graph in the backend to compute the error rate.

\section{Conclusion and Future Work}\label{Conclusion and Future Work}

GPT-based models have been widely implemented for NLP tasks, which demonstrated their extraordinary performance on dealing with language sequence. To address the challenges associated with the heavy reliance on expert knowledge and the time-consuming nature of base Human Error Probability (HEP) estimation, we propose an innovative LLM-based two-stage framework, termed knowledge-driven reliability analysis integrating IDHEAS and LLMs (KRAIL). We have prepared a dataset for experimentation and evaluated the framework’s performance across various configurations. Additionally, its time efficiency relative to manual methods is compared. To further validate the approach, we conducted ablation experiments to examine the effectiveness of the two-stage framework. Finally, a case study was presented to demonstrate the practical applicability and effectiveness of our method. The results show that our KRAIL method limits the time required for base HEP estimation to under 150 seconds, while also achieving high accuracy. To further improve KRAIL’s accuracy and applicability, future work will focus on expanding the knowledge base by incorporating additional data sources and refining the scenario labels. This expansion will allow KRAIL to cover a broader range of operational contexts and error types, ensuring that HEP estimates are more comprehensive and representative of diverse industry needs. Additionally, future developments can aim to fine-tune model parameters and training datasets to improve accuracy.

\backmatter

\section*{Declarations}

\subsection{Funding.} The research was supported by a grant from the National Natural Science Foundation of China (Grant No. T2192933).
\subsection{Author contribution.} Xingyu Xiao: Conceptualization, Methodology, Software, Formal analysis, Data Curation, Visualization, Validation, Writing- Original draft preparation, Writing - Review and Editing. Peng Chen: Software, Methodology. Ben Qi: Methodology. Hongru Zhao: Resources. Jingang Liang: Conceptualization, Resources, Supervision, Writing - Review and Editing, Project administration, Funding acquisition. Jiejuan Tong: Investigation, Supervision, Writing - Review and Editing. Haitao Wang: Supervision, Writing- Reviewing and Editing.

\subsection{Data availability}
The data used for testing will be made publicly available.

\subsection{Declaration of competing interest.}
The authors declare that they have no known competing financial interests or personal relationships that could have appeared to influence the work reported in this paper.

\begin{appendices}

\section{Detailed Outputs of the Case Study in the Multi-Agent Phase}\label{secA1}

\begin{figure}[h]
\centering
\includegraphics[width=1.0\textwidth]{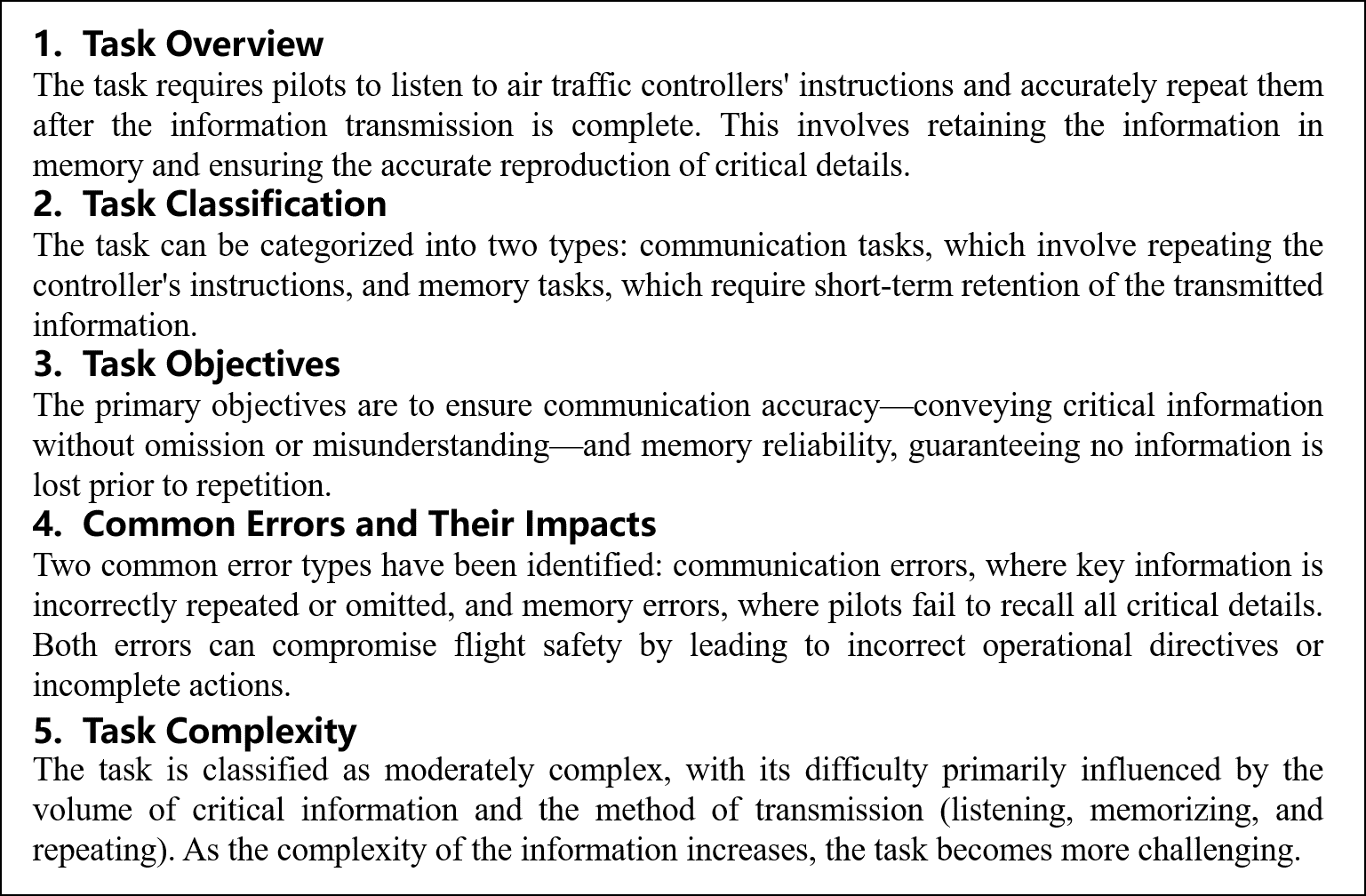}
\caption{Task Analysis Outputs from the Multi-Agent Process in the Case Study.}\label{box1}
\end{figure}

\begin{figure}[H]
\centering
\includegraphics[width=1.0\textwidth]{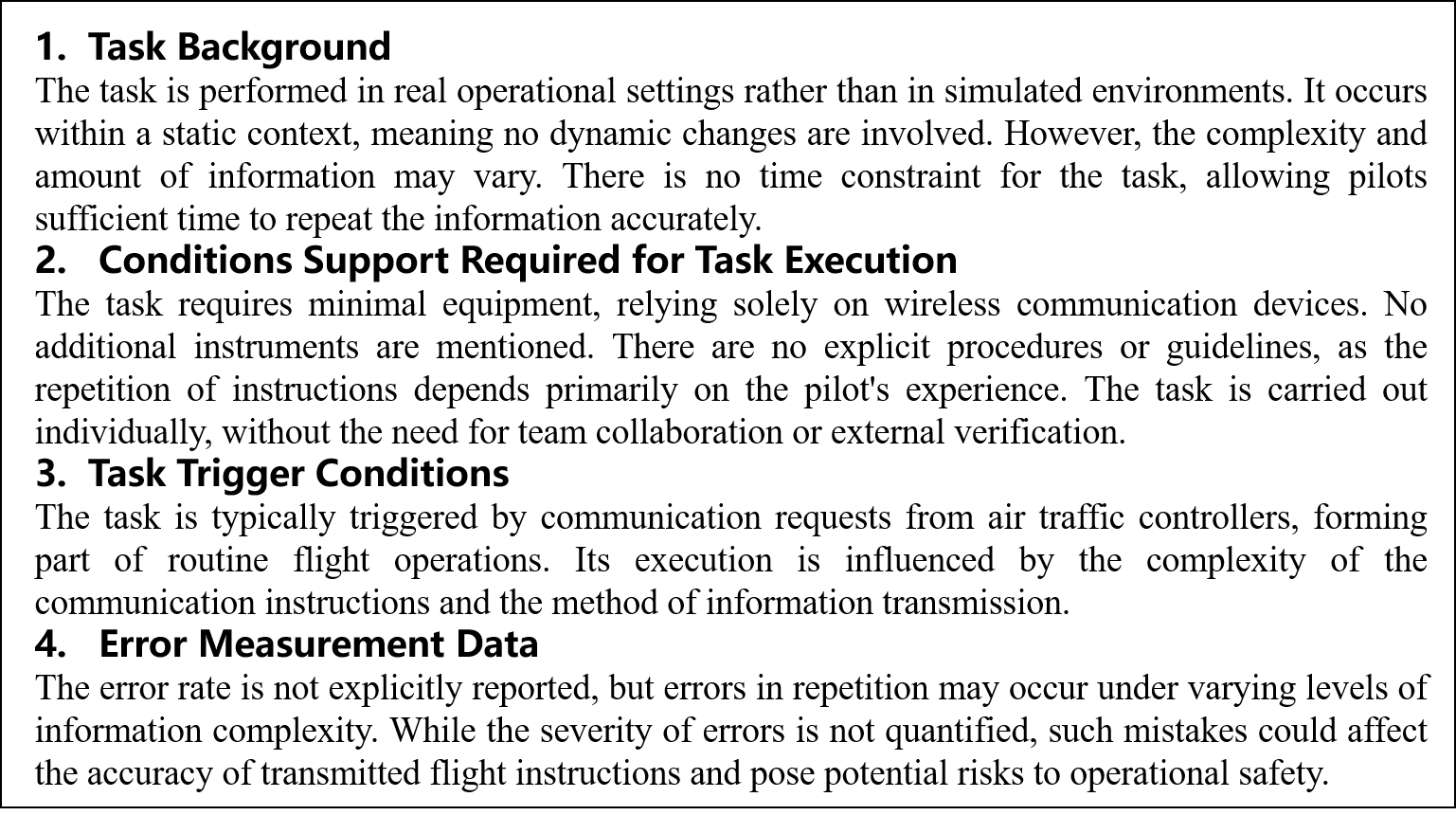}
\caption{Context Analysis Outputs from the Multi-Agent Process in the Case Study.}\label{box2}
\end{figure}

\begin{figure}[H]
\centering
\includegraphics[width=1.0\textwidth]{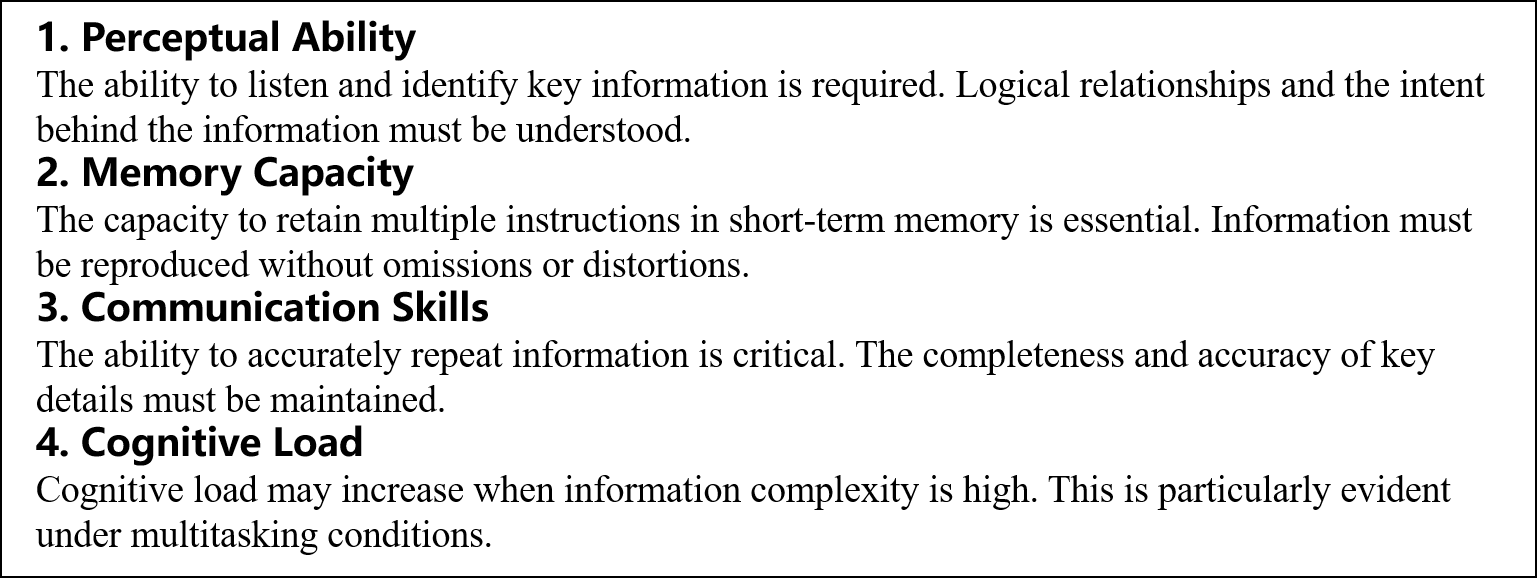}
\caption{Cognitive Activities Analysis Outputs from the Multi-Agent Process in the Case Study.}\label{box3}
\end{figure}

\begin{figure}[H]
\centering
\includegraphics[width=1.0\textwidth]{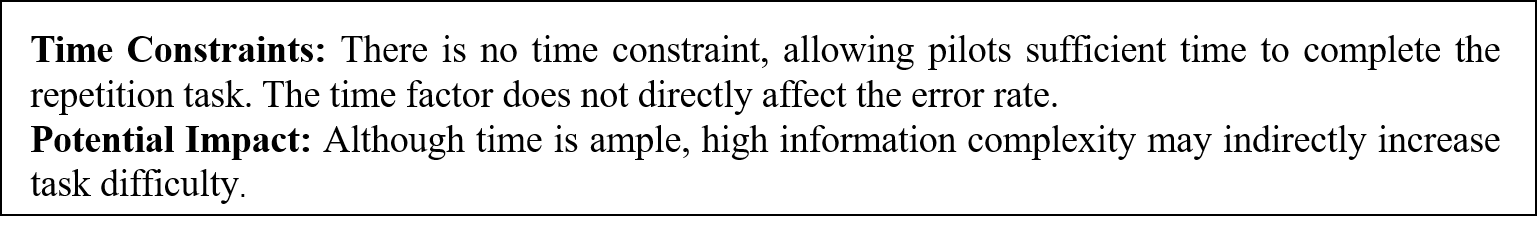}
\caption{Time Constraints Analysis Outputs from the Multi-Agent Process in the Case Study.}\label{box4}
\end{figure}
\end{appendices}

\newpage

% \bibliography{sn-bibliography}% common bib file

\begin{thebibliography}{42}
\providecommand{\natexlab}[1]{#1}
\providecommand{\url}[1]{{#1}}
\providecommand{\urlprefix}{URL }
\providecommand{\doi}[1]{\url{https://doi.org/#1}}
\providecommand{\eprint}[2][]{\url{#2}}
 \bibcommenthead
\bibitem[{Xiao et~al(2024{\natexlab{a}})Xiao, Liang, Tong, and Wang}]{xiao2024emergency}
Xiao X, Liang J, Tong J, et~al (2024{\natexlab{a}}) Emergency decision support techniques for nuclear power plants: Current state, challenges, and future trends. Energies 17(10):2439

\bibitem[{Chang and Xing(2016)}]{chang2016general}
Chang J, Xing J (2016) The general methodology of an integrated human event analysis system (idheas) for human reliability analysis method development. PSAM 13

\bibitem[{Xing et~al(2020)Xing, Chang, and DeJesus}]{xing2020integrated}
Xing J, Chang Y, DeJesus J (2020) Integrated human event analysis system for event and condition assessment (idheas-eca). US Nuclear Regulatory Commission, Washington, DC

\bibitem[{Xing et~al(2021)Xing, CHANG, and SEGARRA}]{xing2021draft}
Xing J, CHANG Y, SEGARRA J (2021) Draft—integrated human event analysis system for human reliability data (idheas-data). RIL-2021-XX

\bibitem[{Cascella et~al(2023)Cascella, Montomoli, Bellini, and Bignami}]{cascella2023evaluating}
Cascella M, Montomoli J, Bellini V, et~al (2023) Evaluating the feasibility of chatgpt in healthcare: an analysis of multiple clinical and research scenarios. Journal of medical systems 47(1):33

\bibitem[{Dubovik et~al(2024)Dubovik, Tishechkin, Kozhevin, Khudorozhkov, Koriagin, Kuvaev, and Altynova}]{dubovik2024advanced}
Dubovik A, Tishechkin D, Kozhevin A, et~al (2024) Advanced llm applications in oil \& gas: Cross-domain implementations. In: 85th EAGE Annual Conference \& Exhibition (including the Workshop Programme), European Association of Geoscientists \& Engineers, pp 1--5

\bibitem[{Xiao et~al(2024{\natexlab{b}})Xiao, Liu, Zuo, Chen, Qi, Liang, and Tong}]{xiao2024text}
Xiao X, Liu S, Zuo Z, et~al (2024{\natexlab{b}}) A text intelligence-based approach for automatic generation of fault trees in nuclear power plants. In: International Conference on Nuclear Engineering, American Society of Mechanical Engineers, p V010T12A004

\bibitem[{Eigner and H{\"a}ndler(2024)}]{eigner2024determinants}
Eigner E, H{\"a}ndler T (2024) Determinants of llm-assisted decision-making. arXiv preprint arXiv:240217385

\bibitem[{Chiang et~al(2024)Chiang, Lu, Li, and Yin}]{chiang2024enhancing}
Chiang CW, Lu Z, Li Z, et~al (2024) Enhancing ai-assisted group decision making through llm-powered devil's advocate. In: Proceedings of the 29th International Conference on Intelligent User Interfaces, pp 103--119

\bibitem[{Ma et~al(2024)Ma, Chen, Wang, Zheng, Peng, Yin, and Ma}]{ma2024towards}
Ma S, Chen Q, Wang X, et~al (2024) Towards human-ai deliberation: Design and evaluation of llm-empowered deliberative ai for ai-assisted decision-making. arXiv preprint arXiv:240316812

\bibitem[{Di~Pasquale et~al(2013)Di~Pasquale, Iannone, Miranda, Riemma et~al}]{di2013overview}
Di~Pasquale V, Iannone R, Miranda S, et~al (2013) An overview of human reliability analysis techniques in manufacturing operations. Operations management 9:978--953

\bibitem[{Cuschieri and Tang(2010)}]{cuschieri2010human}
Cuschieri A, Tang B (2010) Human reliability analysis (hra) techniques and observational clinical hra. Minimally Invasive Therapy \& Allied Technologies 19(1):12--17

\bibitem[{Kirwan(1996)}]{kirwan1996validation}
Kirwan B (1996) The validation of three human reliability quantification techniques—therp, heart and jhedi: Part 1—technique descriptions and validation issues. Applied ergonomics 27(6):359--373

\bibitem[{Hannaman et~al(1985)Hannaman, Spurgin, and Lukic}]{hannaman1985model}
Hannaman G, Spurgin A, Lukic Y (1985) A model for assessing human cognitive reliability in pra studies. In: Conference record for 1985 IEEE third conference on human factors and nuclear safety

\bibitem[{Swain(1987)}]{swain1987accident}
Swain AD (1987) Accident sequence evaluation program: Human reliability analysis procedure. Tech. rep., Sandia National Lab.(SNL-NM), Albuquerque, NM (United States); US Nuclear~…

\bibitem[{Groth et~al(2019)Groth, Smith, and Moradi}]{groth2019hybrid}
Groth KM, Smith R, Moradi R (2019) A hybrid algorithm for developing third generation hra methods using simulator data, causal models, and cognitive science. Reliability Engineering \& System Safety 191:106507


\bibitem[{Hollnagel(1998)}]{hollnagel1998cognitive}
Hollnagel E (1998) Cognitive reliability and error analysis method (CREAM). Elsevier

\bibitem[{Cooper et~al(1996)Cooper, Ramey-Smith, Wreathall, and Parry}]{cooper1996technique}
Cooper SE, Ramey-Smith A, Wreathall J, et~al (1996) A technique for human error analysis (atheana). Tech. rep., Nuclear Regulatory Commission


\bibitem[{Levine et~al(2024)Levine, Al-Douri, Paglioni, Bensi, and Groth}]{levine2024identifying}
Levine CS, Al-Douri A, Paglioni VP, et~al (2024) Identifying human failure events for human reliability analysis: A review of gaps and research opportunities. Reliability Engineering \& System Safety p 109967

\bibitem[{Porthin et~al(2024)Porthin, Podofillini, and Dang}]{porthin2024task}
Porthin M, Podofillini L, Dang VN (2024) Task reliability index for operator performance and failure probability assessment in control room simulators. Reliability Engineering \& System Safety 251:110390

\bibitem[{Park(2021)}]{park2021dynamic}
Park J (2021) Dynamic hra for flex. Tech. rep., Idaho National Lab.(INL), Idaho Falls, ID (United States)

\bibitem[{Reid et~al(2024)Reid, Savinov, Teplyashin, Lepikhin, Lillicrap, Alayrac, Soricut, Lazaridou, Firat, Schrittwieser et~al}]{reid2024gemini}
Reid M, Savinov N, Teplyashin D, et~al (2024) Gemini 1.5: Unlocking multimodal understanding across millions of tokens of context. arXiv preprint arXiv:240305530

\bibitem[{Buehler(2024)}]{buehler2024accelerating}
Buehler MJ (2024) Accelerating scientific discovery with generative knowledge extraction, graph-based representation, and multimodal intelligent graph reasoning. arXiv preprint arXiv:240311996



\bibitem[{Li et~al(2024{\natexlab{a}})Li, Wang, Zhang, Li, Lai, Kang, Ma, and Liu}]{li2024agent}
Li J, Wang S, Zhang M, et~al (2024{\natexlab{a}}) Agent hospital: A simulacrum of hospital with evolvable medical agents. arXiv preprint arXiv:240502957

\bibitem[{Oettinger et~al(1990)Oettinger, Schatz, Gorka, and Baltimore}]{oettinger1990rag}
Oettinger MA, Schatz DG, Gorka C, et~al (1990) Rag-1 and rag-2, adjacent genes that synergistically activate v (d) j recombination. Science 248(4962):1517--1523

\bibitem[{Wang et~al(2024)Wang, Long, Xiao, Cai, Wu, Meng, Wang, and Zhou}]{wang2024biorag}
Wang C, Long Q, Xiao M, et~al (2024) Biorag: A rag-llm framework for biological question reasoning. arXiv preprint arXiv:240801107

\bibitem[{Li et~al(2024{\natexlab{b}})Li, Yuan, and Zhang}]{li2024enhancing}
Li J, Yuan Y, Zhang Z (2024{\natexlab{b}}) Enhancing llm factual accuracy with rag to counter hallucinations: A case study on domain-specific queries in private knowledge-bases. arXiv preprint arXiv:240310446

\bibitem[{Xing and Chang(2018)}]{xing2018use}
Xing J, Chang Y (2018) Use of idheas general methodology to incorporate human performance data for estimation of human error probabilities. In: 14th International Conference on Probabilistic Safety Assessment and Management (PSAM 14), Los Angeles, CA, US, pp 16--21



\bibitem[{Ahmed and Devanbu(2022)}]{ahmed2022few}
Ahmed T, Devanbu P (2022) Few-shot training llms for project-specific code-summarization. In: Proceedings of the 37th IEEE/ACM International Conference on Automated Software Engineering, pp 1--5


\bibitem[{Cohen et~al(2012)Cohen, Smetzer, Westphal, Comden, and Horn}]{cohen2012risk}
Cohen MR, Smetzer JL, Westphal JE, et~al (2012) Risk models to improve safety of dispensing high-alert medications in community pharmacies. Journal of the American Pharmacists Association 52(5):584--602

\bibitem[{McGuirl and Sarter(2006)}]{mcguirl2006supporting}
McGuirl JM, Sarter NB (2006) Supporting trust calibration and the effective use of decision aids by presenting dynamic system confidence information. Human factors 48(4):656--665

\bibitem[{Xing(2017)}]{xing2017integrated}
Xing J (2017) An integrated human event analysis system (IDHEAS) for nuclear power plant internal events at-power application. US Nuclear Regulatory Commission, Office of Nuclear Regulatory Research (RES)

\bibitem[{Preischl and Hellmich(2013)}]{preischl2013human}
Preischl W, Hellmich M (2013) Human error probabilities from operational experience of german nuclear power plants. Reliability Engineering \& System Safety 109:150--159

\bibitem[{Preischl and Hellmich(2016)}]{preischl2016human}
Preischl W, Hellmich M (2016) Human error probabilities from operational experience of german nuclear power plants, part ii. Reliability Engineering \& System Safety 148:44--56



\bibitem[{Aalipour et~al(2016)Aalipour, Ayele, and Barabadi}]{aalipour2016human}
Aalipour M, Ayele YZ, Barabadi A (2016) Human reliability assessment (hra) in maintenance of production process: a case study. International Journal of System Assurance Engineering and Management 7:229--238

\bibitem[{Lee et~al(2002)Lee, McGehee, Brown, and Reyes}]{lee2002collision}
Lee JD, McGehee DV, Brown TL, et~al (2002) Collision warning timing, driver distraction, and driver response to imminent rear-end collisions in a high-fidelity driving simulator. Human factors 44(2):314--334

\bibitem[{Kurokawa et~al(2024)Kurokawa, Ohizumi, Kanzawa, Kurokawa, Sonoda, Nakamura, Kiguchi, Gonoi, and Abe}]{kurokawa2024diagnostic}
Kurokawa R, Ohizumi Y, Kanzawa J, et~al (2024) Diagnostic performances of claude 3 opus and claude 3.5 sonnet from patient history and key images in radiology’s “diagnosis please” cases. Japanese Journal of Radiology pp 1--4


\bibitem[{Miller(2013)}]{miller2013graph}
Miller JJ (2013) Graph database applications and concepts with neo4j. In: Proceedings of the southern association for information systems conference, Atlanta, GA, USA, pp 141--147




\bibitem[{Campochiaro et~al(2009)Campochiaro, Casatta, Cremonesi, and Turrin}]{campochiaro2009metrics}
Campochiaro E, Casatta R, Cremonesi P, et~al (2009) Do metrics make recommender algorithms? In: 2009 International Conference on Advanced Information Networking and Applications Workshops, IEEE, pp 648--653


\bibitem[{Mooney et~al(1993)Mooney, Duval, and Duvall}]{mooney1993bootstrapping}
Mooney CZ, Duval RD, Duvall R (1993) Bootstrapping: A nonparametric approach to statistical inference. 95, sage

\bibitem[{Kim(2015)}]{kim2015t}
Kim TK (2015) T test as a parametric statistic. Korean journal of anesthesiology 68(6):540--546




\bibitem[{Gertman et~al(2005)Gertman, Blackman, Marble, Byers, Smith et~al}]{gertman2005spar}
Gertman D, Blackman H, Marble J, et~al (2005) The spar-h human reliability analysis method. US Nuclear Regulatory Commission 230(4):35































\end{thebibliography}
% %% if required, the content of .bbl file can be included here once bbl is generated

\end{document}